\documentclass{article}

\usepackage{arxiv}

\usepackage[utf8]{inputenc} 
\usepackage[T1]{fontenc}    
\usepackage{hyperref}       
\usepackage{url}            
\usepackage{booktabs}       
\usepackage{amsfonts}       
\usepackage{nicefrac}       
\usepackage{microtype}      
\usepackage{lipsum}
\usepackage{graphicx}
\usepackage{amsmath}
\usepackage[ruled,vlined]{algorithm2e}
\usepackage{pifont}
\newcommand{\cmark}{\ding{51}}%
\newcommand{\xmark}{\ding{55}}%

\title{L2P: Learning to Place for \\ Estimating Heavy-Tailed Distributed Outcomes}

\author{
 Xindi Wang \\
  Network Science Institute\\
  Northeastern University\\
  Boston, BA 02115 \\
  \texttt{wang.xind@northeastern.edu} \\
   \And
 Onur Varol \\
  Faculty of Engineering and Natural Sciences\\
  Sabancı University, \\
  Istanbul, Turkey 34956\\
  \texttt{onur.varol@sabanciuniv.edu} \\
  \And
 Tina Eliassi-Rad \\
  Network Science Institute\\
  Khoury College of Computer Sciences\\
  Northeastern University\\
  Boston, BA 02115 \\
  \texttt{eliassi@northeastern.edu} \\
}

\begin{document}
\maketitle

\begin{abstract}
Many real-world prediction tasks have outcome variables that have characteristic heavy-tail distributions. Examples include copies of books sold, auction prices of art pieces, demand for commodities in warehouses, etc. By learning heavy-tailed distributions, ``big and rare'' instances (e.g., the best-sellers) will have accurate predictions. Most existing approaches are not dedicated to learning heavy-tailed distribution; thus, they heavily under-predict such instances. To tackle this problem, we introduce \emph{Learning to Place} (\texttt{L2P}), which exploits the pairwise relationships between instances for learning. In its training phase, \texttt{L2P} learns a pairwise preference classifier: \texttt{is instance A $>$ instance B?} In its placing phase, \texttt{L2P} obtains a prediction by placing the new instance among the known instances. Based on its placement, the new instance is then assigned a value for its outcome variable. Experiments on real data show that  \texttt{L2P} outperforms competing approaches in terms of accuracy and ability to reproduce heavy-tailed outcome distribution. In addition, \texttt{L2P} provides an interpretable model by placing each predicted instance in relation to its comparable neighbors. Interpretable models are highly desirable when lives and treasure are at stake.
\end{abstract}

\section{Introduction}

Heavy-tailed distributions are prevalent in real world data. By heavy-tailed, we mean a variable whose distribution has a heavier tail than the exponential distribution. Many real-world applications involve predicting heavy-tailed distributed outcomes. For example, publishers want to predict a book's sales number before its publication in order to decide the advance for the author, effort in advertisement, copies printed, etc~\cite{wang2019success}. Galleries are interested in an artist's selling potential to decide whether to represent the artist or not. For inventory planning, warehouses and shops would like to know what number of each item to keep in storage~\cite{syntetos2009forecasting}. All of these real-world applications involve heavy-tailed distributed outcomes -- book sales, art auction prices, and demands for items.

The challenge for predicting heavy-tailed outcomes lies at the tail of the distributions, i.e., the ``big and rare'' instances such as best-sellers, high-selling art pieces and items with huge demands. Those instances are usually the ones that attract the most interests and create the most market values. Traditional approaches tend to under-predict the rare instances at the tail of the distribution. The limiting factor for prediction performance is the insufficient amount of training data on the rare instances. Approaches tackling class imbalance problem, such as over-sampling training instances~\cite{chawla2002smote}, adjusting weights, and adding extra constraints~\cite{maalouf2018logistic,ertekin2007learning,schubach2017imbalance,king2001logistic} do not properly address the aforementioned problem, since those approaches assume homogeneously distributed groups with proportionally different sizes. In addition, defining distinct groups on a dataset with heavy-tailed outcomes is not trivial since the  distribution is continuous. Therefore, predicting the values of heavy-tailed variables is not merely the class imbalance problem; instead it is the problem of learning a heavy-tailed distribution. 

We present an approach called \emph{Learning to Place} (\texttt{L2P}) to estimate heavy-tailed outcomes and define performance measures for heavy-tailed target variables prediction. \texttt{L2P} learns to estimate a heavy-tailed distribution by first learning pairwise preferences between the instances and then placing the new instance within the known instances and assigning an outcome value. Our contributions are as follows:

\begin{enumerate}
\item We introduce \emph{Learning to Place} (\texttt{L2P}) to estimate heavy-tailed outcomes by (i) learning from pairwise relationships between instances and (ii) placing new instances among the training data and predicting outcome values based on those placements. 
\item We present appropriate statistical metrics for measuring the performance for learning  heavy-tailed distributions.
\item In an exhaustive empirical study on real-world data, we demonstrate that \texttt{L2P} is robust and consistently outperforms various competing approaches across diverse real-world datasets.
\item Through case studies, we demonstrate that \texttt{L2P} not only provides accurate predictions but also is interpretable.
\end{enumerate}


The outline of the paper is as follows. Section 2 presents our proposed method. Section 3 describes our experiments. Section 4 explains how our method produces models that are interpretable. Section 5 contains related work and a discussion of our findings. The paper concludes in Section 6. 

\section{Proposed Method: Learning to Place (L2P)}
\texttt{L2P} takes as input a data matrix where the rows are data instances (e.g., books) and the columns are features that describe each instance (e.g., author, publisher, \texttt{etc}).  Each data instance also has a value for the predefined target variable (e.g., copies of book sold).  \texttt{L2P} learns to map each instance's feature vector to the value for its target variable, which is the standard supervised learning setup.  However, the challenges that \texttt{L2P} addresses are as follows.  First, it learns the heavy-tailed distribution of the target variable; and thus it does not under-predict the ``big and rare'' instances. Second, it generates an interpretable model for the human end-user (e.g., a publisher).

\begin{figure}[!ht]
\centering
  \includegraphics[width=\linewidth]{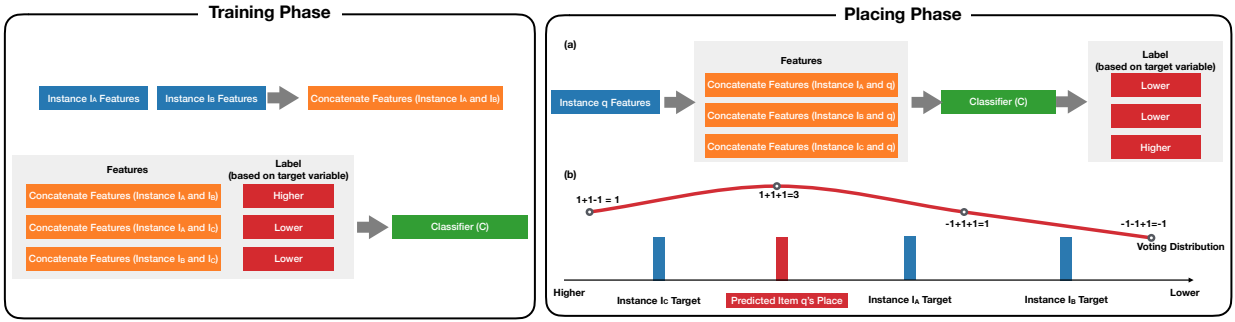}
  \caption{\textbf{Overview of the \texttt{L2P} Algorithm.} In the training phase, \texttt{L2P} learns a classifier $C$ on the pairwise-relationship between each pair of training instances. In the placing phase, \texttt{L2P} applies the classifier $C$ to obtain the pairwise relationships between unplaced instance $q$ and all of the training instances. Then, each instance in the training set contributes to the placement of unplaced instance $q$ by voting on bins to its left or to its right depending on the predicted relation between instances. The mid-value of the bin with the highest vote is assigned as the prediction.
  }
  \label{fig:schema_l2p}
\end{figure}

Figure~\ref{fig:schema_l2p} describes the training and placing phases for \texttt{L2P}.  In the training phase, \texttt{L2P} learns a pairwise-relationship classifier, which predicts whether the target variable for an instance A (\emph{$I_{A}$}) is greater (or less) than another instance B (\emph{$I_{B}$}).  To predict outcomes in the placing phase, the unplaced instance is compared with each training instance using the model learned in the training phase, generating pairwise-relationships between the unplaced instance and the training instances.  These pairwise-relationships are then used as ``votes'' to predict the target outcome of the new instance.

\noindent\textbf{Training Phase.} For each pair of instances $i$ and $j$ with feature vector $f_i$ and $f_j$, \texttt{L2P} concatenates the two feature vectors $X_{ij} = [f_i, f_j]$.\footnote{We also tried L2 norm feature difference and observe that the model's performance is similar.} If $i$'s target variable is greater than $j$'s, then $y_{ij} =1$; otherwise, $y_{ij} = -1$ (ties are ignored in the training phase). Formally, denoting with $t_i$ the target variable for instance $i$ and with $S$ the set of instances in the training set, \texttt{L2P} generates the following training data:
\begin{equation}
X_{ij} = [f_i, f_j], \text{ for each} (i,j) \in S \times S, i\neq j\ , t_i \neq t_j,
\end{equation}
\begin{equation}
y_{ij} = \begin{cases}
1,&t_i>t_j\\
-1, &t_i<t_j
\end{cases}.
\end{equation}
Then a classifier $C$ is trained on the training data $X_{ij}$ and labels $y_{ij}$.
\footnote{Training on a single $(i,j)$ pair or training on symmetric pairs (i.e., including both $(i,j)$ and $(j,i)$) does not produce significant differences in the model's performance in our experiments.}
It is important to note that the trained classifier may produce conflicting results: for example, $I_{A} < I_{B}$ and $I_{B} < I_{C}$ but $I_{C} < I_{A}$. In the Experiments section, we demonstrate the robustness of \texttt{L2P} to such conflicts rooted from the pairwise classification error.

\noindent\textbf{Placing Phase.} The placing phase consists of two stages. In Stage I, for each unplaced instance $q$, \texttt{L2P} obtains $X_{iq} = [f_i, f_q], \text{ for each } i \in S$ (recall $S$ is the training set). Then, \texttt{L2P} applies the classifier $C$ on $X_{iq}$ to get the predicted pairwise relationship between the test instance $q$ and all training instances ($\hat{y}_{iq} = C(X_{iq})$). In Stage II, \texttt{L2P} treats each training instance as a ``voter''. Training instances (voters) are sorted by their target variables in descending order, dividing the target variable axis into bins. If $\hat{y}_{iq} = 1$, bins on the right of $t_i$ will obtain an \emph{upvote} (+1) and bins on the left of $t_i$ will obtain a \emph{downvote} (-1) and vice versa for $\hat{y}_{iq} = -1$. After the voting process, \texttt{L2P} takes the bin with the most ``votes'' as the predicted bin for test instance $q$, and obtains the prediction $\hat{t}_q$ as the midpoint of this bin.

\subsection{Theoretical Analysis of Voting Process}
\texttt{L2P}'s voting process can be viewed as the maximum likelihood estimation (MLE) of the optimal placement of an instance based on the pairwise relationships. Given the test instance $q$, our goal is to find its optimal bin $m$. For any bin $b$, we have: 
$P(b|q) \propto P(q|b)\times P(b)$. Since each train instance $i$ contributes to $P(q|b)$, we have $P(q|b) = \frac{1}{Z}\sum_{i \in S_{train}} P_i(q|b)$, where $P_i(q|b)$ is the conditional probability of test instance $q$ placing in the given bin $b$ based on its pairwise relationship with training instance $i$; and $Z$ is the normalization factor, $Z = \sum_b \sum_i P_i(q|b)$. \texttt{L2P} assigns two probabilities to each pair of training instance $i$ and test instance $q$:  $p_i^l(q)$ and $p_i^r(q)$, denoting the probability that the test instance $q$ is smaller than (i.e., to the left of) or larger than (i.e., to the right of) training instance $i$ respectively, where $p_i^l(q) + p_i^r(q) = 1$.\footnote{We are not voting by probability, but imposing a step function after obtaining the probability: i.e., below a probability threshold, vote -1 and above vote 1. This does not influence the big picture of the proof.}
Let $R_b^i \in \{l, r\}$ be the region defined by training instance $i$ for bin $b$, and $|R_b^i|$ as the number of bins in this region. We know $P_i(q|b) = p_i^{R_b^i}(q)/|R_b^i|$ assuming the test instance is equally probable to fall in each bin in region $R_m^i$.\footnote{This assumption is reflected in our voting strategy: downvote on one side and upvote on the other side.}
Therefore, the optimal bin is $m = \text{argmax}_b \frac{1}{Z}\sum_{i \in S} (p_i^{R_b^i}(q)/|R_b^i|)$. We observe that $p_i^{R_b^i}(q)/|R_b^i|$ is the ``votes'' the training instance $i$ gives to bin $b$ for test instance $q$, therefore the optimal bin $m$ is the one with the most ``votes''.\footnote{Notice that by using upvotes (+1) and downvotes (-1) in our approach, we are basically standardizing $p_i^l(q)$ and $p_i^r(q)$.}

\begin{algorithm}[t]
    \SetAlgoLined
    \SetNoFillComment
    \DontPrintSemicolon
    \KwIn{Training data $S=(F,t)$}
    \KwOut{Pairwise relationship classifier $C$}
    \tcp*[l]{Feature matrix}
    $X = [\ ]$;\\
    \tcp*[l]{Label vector}
    $y = [\ ]$;\\
    \For{$i\leftarrow 1$ \KwTo $|S|$}{
        \For{$j\leftarrow i+1$ \KwTo $|S|$}{
            $X$.append([$f_i$, $f_j$]);\\
            \uIf{$t_i > t_j$}{
                $y$.append(1)\;
            }
            \uElseIf{$t_i < t_j$}
                {$y$.append(-1)\;
            }
        }
    }
    C.train($X$, $y$);\\
    \Return{C}
    \caption{\texttt{L2P}'s Training Algorithm}
    \label{alg:l2ptrain}
  \end{algorithm}

\begin{algorithm}[t]
        \SetAlgoLined
        \SetNoFillComment
        \DontPrintSemicolon
        \KwIn{Classifier $C$, Training data $S$=($F$, $t$), Unplaced instance $q$ represented by its features $f_q$}
        \KwOut{$t_q$ = predicted value for test instance $q$}
        $B = [\ ]$;\\
        \tcp*[l]{Unique target values, highest to lowest}
        bins = sort(unique($t$));\\
        \For{$i \leftarrow 1$ \KwTo $|\text{bins}|$}{
            B[i] = 0\;
        }
        \For{$i \leftarrow 1$ \KwTo $|F|$}{
            $\hat{y}_{iq} = C.predict([f_i, f_q])$\;
            \For{$j \leftarrow 1$ \KwTo
            BinEdgeIndex($t_i$) - 1}{
                B[j] -= $\hat{y}_{iq}$; 
            }
            \For{$j \leftarrow$ BinEdgeIndex($t_i$) \KwTo $|B|$}{
                B[j] += $\hat{y}_{iq}$; 
            }
        }
        b = GetHighestBin(B);\\
        $t_q$ = Mean(bins[b-1], bins[b]);\\
        \Return{$t_q$}
        \\ 
         \caption{\texttt{L2P}'s Placing Algorithm}
         \label{alg:l2ptest}
        \end{algorithm}

\texttt{L2P} can incorporate any method that takes pairwise preferences and learns to place a test instance among the training instances. Specifically, we examined  SpringRank~\cite{de2018physical}, FAS-PIVOT~\cite{ailon2008aggregating} and tournament graph related heuristics~\cite{cohen1998learning}. We found that the performances of these approaches are quite similar to voting.  However, voting -- with its linear runtime complexity -- is the most computationally efficient method among them.

\subsection{Complexity Analysis}
Suppose $n$ is the number of instances in the training set. The vanilla training phase for \texttt{L2P} learns pairwise relationships among all pairs in the training set, leading to a $O(n^2)$ runtime complexity, which is computationally prohibitive for large datasets even though the training phase is an offline process and can be easily parallelized. 
 Thus, we implemented an efficient approach based on the intuition that it is easier for a classifier to learn the pairwise relationship between instances that are far apart than instances that are near each other. Specifically, we define two parameters: $n_s$ denoting the number of samples to compare with for each training instance and $k$ denoting the number of instances that are considered near to each training instance. For comparison to each training instance $i$, \texttt{L2P}'s efficient training phase algorithm (1) takes all $k$ near instances of $i$ and (2) uniformly at random samples $n_s - k$ non-near instances of $i$. The nearness of two instances is measured by the difference in their target values. Our experiments with the efficient implementation of \texttt{L2P} lead to similar AUC scores for the overall prediction task, but reduced the runtime complexity to $O(n_s\times n)$, where $n_s << n$.  For instance, $n_s$ is 20\% of $n$. \texttt{L2P}'s placing phase has a complexity of $O(n)$ for each new (i.e., test) instance.

\section{Experiments}
In this section, we describe the data used in our experiments, the baseline and competing approaches, experimental methodology, evaluation metrics, and results.

\subsection{Datasets}
We present results on the following real-world applications:

\noindent\textbf{Book sales}: This dataset consists of information about all print nonfiction and fiction books published in the United States in 2015, including features about authors publication history, book summary, authors popularity (details of features see~\cite{wang2019success}). The goal is to predict the one year book sales using features prior to the book's publication. We separate nonfiction books and fiction books in the experiment.

\noindent\textbf{Art auctions}: This dataset combines information on artists exhibits, auction sales, and primary market quotes. 
It was previously used to quantify success of artists based on their trajectories of exhibitions~\cite{fraiberger2018quantifying}.
We select 7,764 paintings using vertical logarithmic binning \cite{henderson2011s}. The features includes previous exhibition records (number of exhibitions, number of exhibitions at different grade), sales records (number of art pieces sold, various statistics of price of previous sold pieces), career length and medium information (full feature list see Supplementary Information Table 1). The prediction task is to predict auction price of an art piece based on artists' previous sale and exhibition history.

\begin{figure}
    \centering
    \includegraphics[width=0.4\columnwidth]{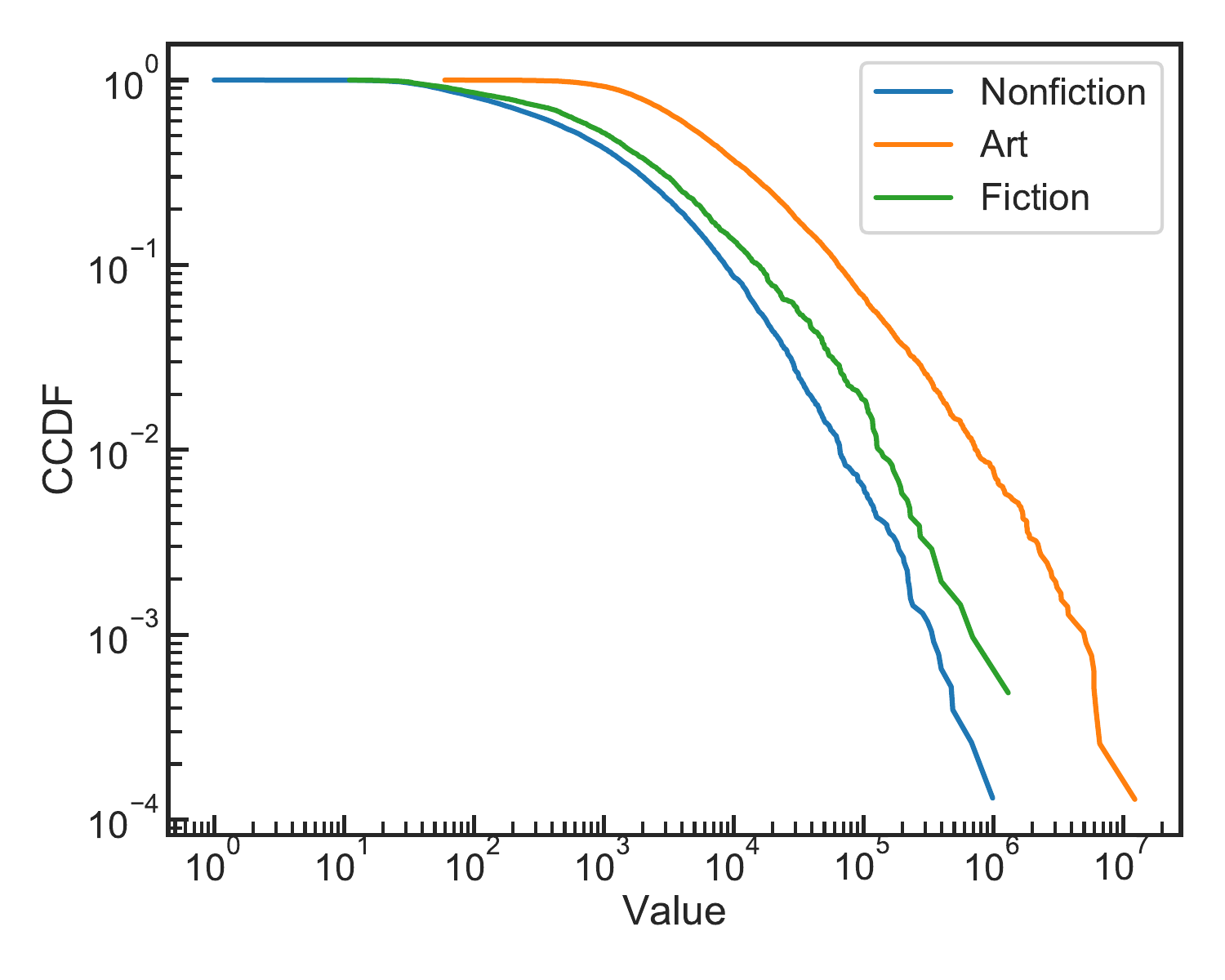}
    \caption{\textbf{Our target variables, book sales and art auctions, have heavy-tailed distributions.}}
    \label{fig:distributions}
  \end{figure}

Table~\ref{tab:datasets} provides the summary statistics of these datasets. Specifically, we calculate the kurtosis for each target variable. Kurtosis measures the ``tailedness'' of the probability distribution. The kurtosis of any univariate normal distribution is 3, and the higher the kurtosis is, the heavier the tails. Complementary cumulative function (CCDF) of real outcomes are shown in Fig.~\ref{fig:distributions}.

\begin{table}[ht]
    \centering
    \begin{tabular}{ccccc}
        \hline
        Name & Instances & Features & Outcome & Kurtosis\\
        \hline
        Nonfiction & 7,641 & 55 & Sales & 619.73\\
        Art & 7,764 & 21 & Price &   597.82\\
        Fiction & 2,061 & 55 & Sales & 443.08\\
        \hline
    \end{tabular}
    \caption{\textbf{Dataset statistics, sorted by Kurtosis values.} Recall that  Kurtosis measures the ``tailedness'' of a real-valued random variable's probability distribution.}
    \label{tab:datasets}
\end{table}

\subsection{Baseline and Competing Approaches}
\label{sec:othermethods}

To compare the predictive capabilities of \texttt{L2P}, we experiment with these baseline approaches from the literature.

\noindent\textbf{K-nearest neighbors (kNN).} The prediction is obtained through local interpolation of nearest neighbors in the training set.

\noindent\textbf{Kernel regression (KR).} We employ Ridge regression with RBF kernels to estimate the non-linear relationship between variables.

\noindent\textbf{Heavy-tailed linear regression (HLR).} Hsu and Sabato~\cite{hsu2014heavy} proposed a heavy-tailed regression model, in which a median-of-means technique is utilized.  They proved that for a $d$-dimensional estimator, a random sample of size $\tilde{O}(d\log(1/\delta))$ is sufficient to obtain a constant factor approximation to the optimal loss with probability $1-\delta$. However, this approach is not able to capture non-linear relationships, which exist in our setting.

\noindent\textbf{Neural networks (NN).} We use a multi-layer perceptron regressor model. Neural networks sacrifice interpretability for predictive accuracy.

\noindent\textbf{XGBoost (XGB).} We use the implementation of gradient boosting optimized on various tasks such as regression and ranking~\cite{chen2016xgboost}.


\noindent\textbf{LambdaMART.} We compare with the well-known learning-to-rank algorithm LambdaMART\footnote{We also experimented with RankSVM\cite{joachims2002optimizing} but this algorithm did not converged.}~\cite{burges2010ranknet}. LambdaMART combines LambdaRank~\cite{burges2007learning} and MART (Multiple Additive Regression Trees)~\cite{friedman2001greedy}. While MART uses gradient boosted decision trees for prediction tasks, LambdaMART uses gradient boosted decision trees using a cost function derived from LambdaRank for solving a ranking task. Here we choose to optimize LambdaMART based on ranking AUC, which is an appropriate metric for our task. LambdaMART is designed to predict the ranking of a list of items; here we predict the instance by: (1) using LambdaMART to get the ranking of the new instance plus the training instances and (2) predicting the value as the midpoint of actual values of adjacent ranked instances.\footnote{LambdaMART code is obtained from \url{https://github.com/jma127/pyltr}.}

\noindent\textbf{Random (RDM).} We shuffle actual outcomes at random and assign those random outcomes as predictions.

\subsection{Experimental Setup and Evaluation Metrics}

We impose standard scaling on all the columns of the data matrix and the target variable.  For all competing methods, we first take the logarithm of the entries in the data matrix and the target variable, and then do standard scaling.

For all methods, we employ 5-fold stratified cross-validation to estimate confidence of model performance. For all baseline models, we tune the model parameters to near-optimal performance (parameters listed in Supplementary Information Table 2). For \texttt{L2P}, however, we use the \texttt{scikit-learn} default parameters of random forest classifier, with 100 trees and Gini impurity as split criteria.

Under our problem definition, the model with the best performance will (1) reproduce heavy-tailed outcome distribution and (2) predict instances accurately especially the ``big and rare instances''. Traditional regression metrics are not of good fit in this problem setting. For example $R^2$ has the assumption of normally distributed error, which is not the case for heavy-tailed distributions; root mean square error (RMSE) will be dominated by the errors on the high end since the values on high-end are extreme. Instead, we use the following metrics that are more appropriate under our circumstances for evaluation:

\noindent\textbf{Quantile-quantile (Q-Q) plot.} Q-Q plot can visually present the deviations between true and predicted target variable distributions. A model who can reproduce the outcome distribution should produce curves closer to $y=x$ line. We can also investigate when predicted quantiles deviate from this line. 

\noindent\textbf{Kolmogorov-Smirnov statistic (KS) and Earth mover distance (EMD).} KS statistic and EMD are two commonly used measure on the distance between two underlying probability distributions. Smaller KS and EMD indicates higher similarity between distributions; and in our analysis, it indicates a better prediction of the underlying distribution.

\noindent\textbf{Receiver operating characteristic (ROC).}
We calculate true-positive and false-positive rates in order to compute the ROC curve and the area under the ROC curve (a.k.a.~AUC score). We adapt AUC calculation to regression setting by calculating the true-positive and false-positive rates at different thresholds (all possible actual values). 

\textbf{True positive rate at threshold.} The calculation of true-positive rate (TPR) at threshold follows the \emph{recall@k} measure used in information retrieval literature. Here, we measure the fraction of instances with true value ($y$) higher than threshold $t$ that also have predicted values ($\hat{y}$) higher than $t$ for calculating TPR. We followed similar approach to compute FPR as in Eq.\ref{eq:tpr}.


\begin{equation}
TPR@t = \frac{\vert \{\hat{y}_i \geq t,  y_i \geq t\}\vert}{\vert \{y_i \geq t\}\vert }
\text{\ \ \ \ \ \ }
FPR@t = \frac{\vert \{\hat{y}_i \geq t, y_i < t\}\vert}{\vert \{y_i < t\}\vert }
\label{eq:tpr}
\end{equation}

For various thresholds, we compute corresponding TPR and FPR scores to create ROC curve.
Similar to traditional ROC curves, a better performing method would have a curve that is simultaneously improving both TPRs and FPRs, leading to a perfect score of AUC = 1. A random model leads to an AUC performance of 0.5 with the corresponding ROC curve being the 45-degree line indicating that TPR and FPR are equal for various thresholds.

It is important to note that \textbf{each individual measure alone is \texttt{not} sufficient to judge the goodness of a model}. KS, EMD and Q-Q plot are measuring the reproducibility of the heavy-tailed distribution, but are not able to measure the prediction accuracy for each instance. Downside of comparing distribution is that the random prediction will end up having the perfect Q-Q plot and KS = 0, EMD = 0. AUC is measuring the accuracy of the prediction, but does not address model's ability to reproduce the distribution. We aim to harness benefits of different measures to assess quality of our models.

    

    

\subsection{Experimental Results}
Next we detail performance comparisons and robustness analysis.

\begin{figure*}
\centering

  \includegraphics[height=0.295\linewidth]{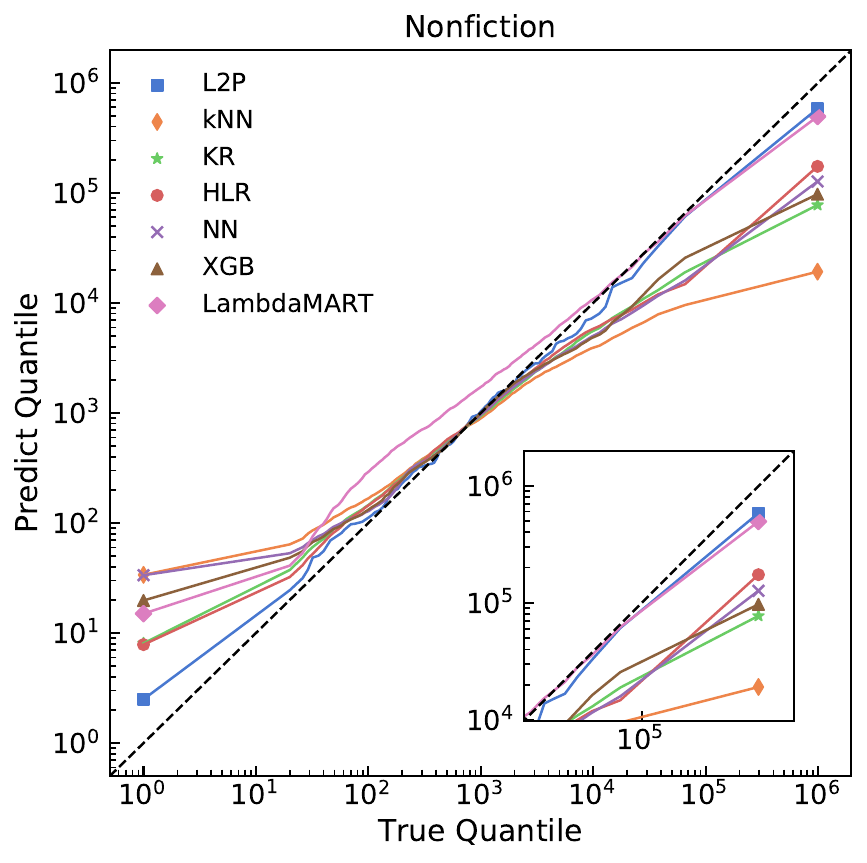}
  \includegraphics[height=0.30\linewidth]{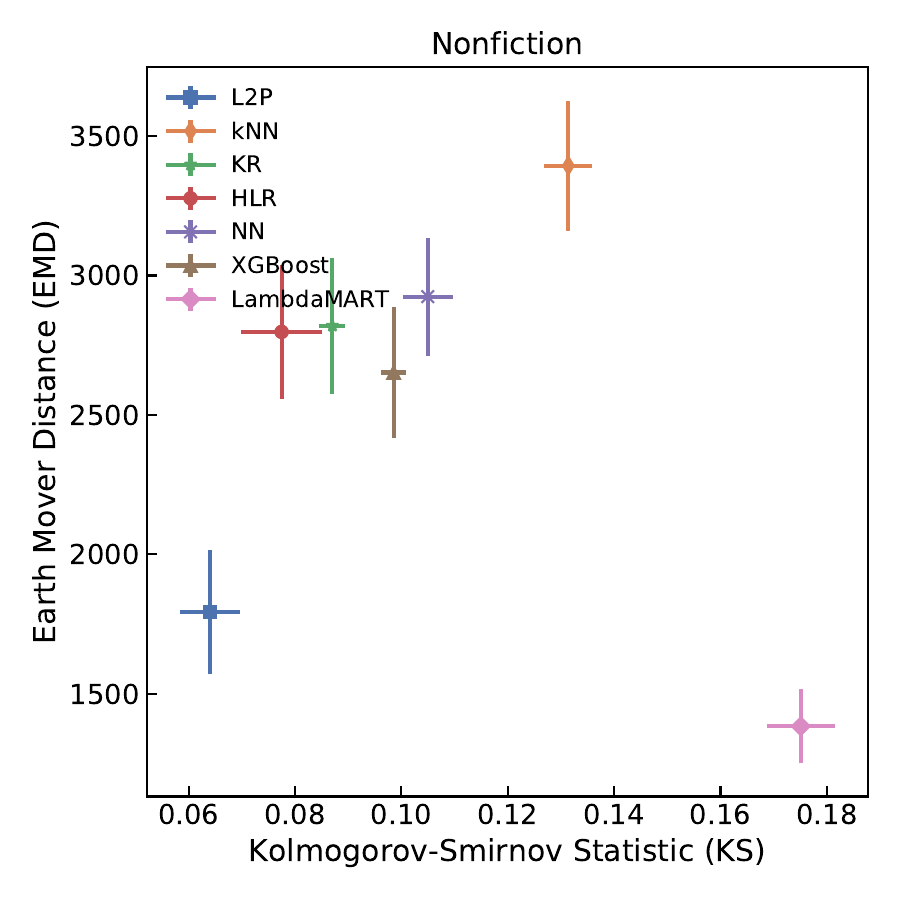}
  \includegraphics[height=0.30\linewidth]{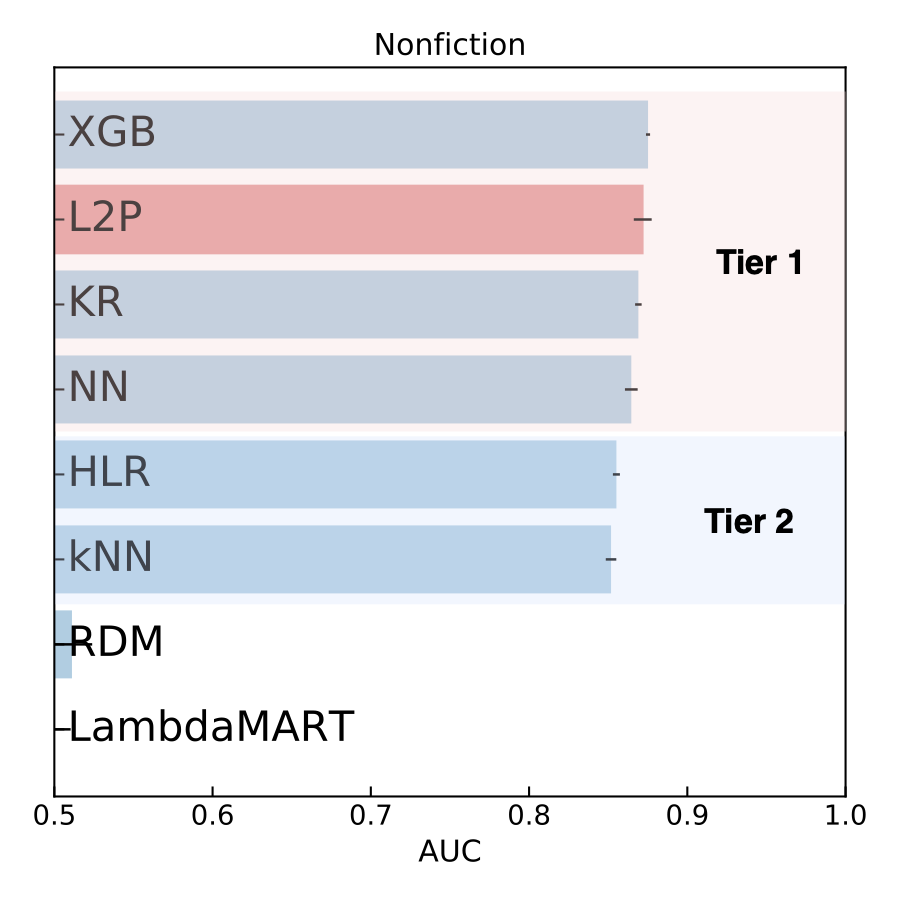} \\

  \includegraphics[height=0.295\linewidth]{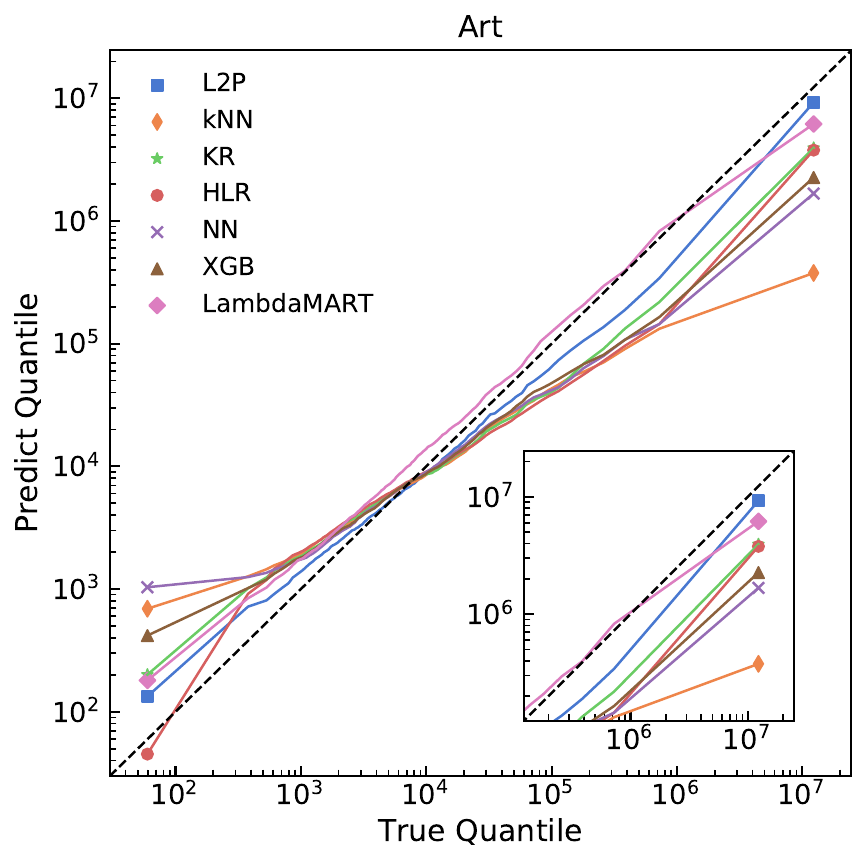}
  \includegraphics[height=0.30\linewidth]{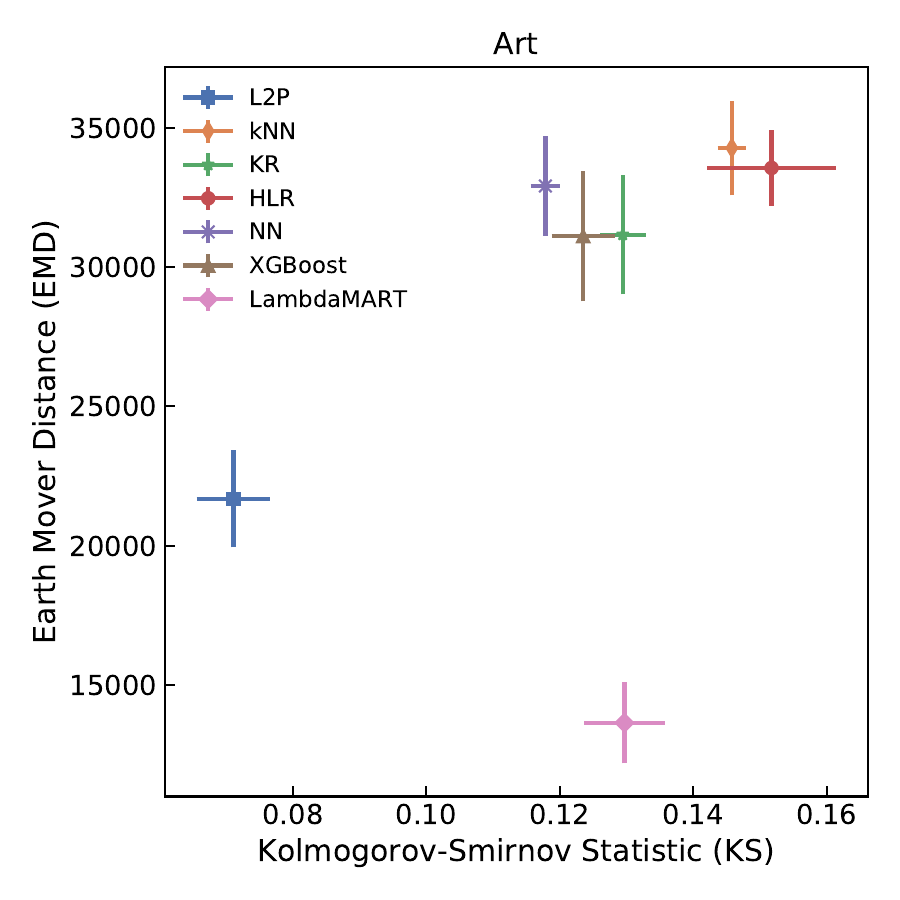}
  \includegraphics[height=0.30\linewidth]{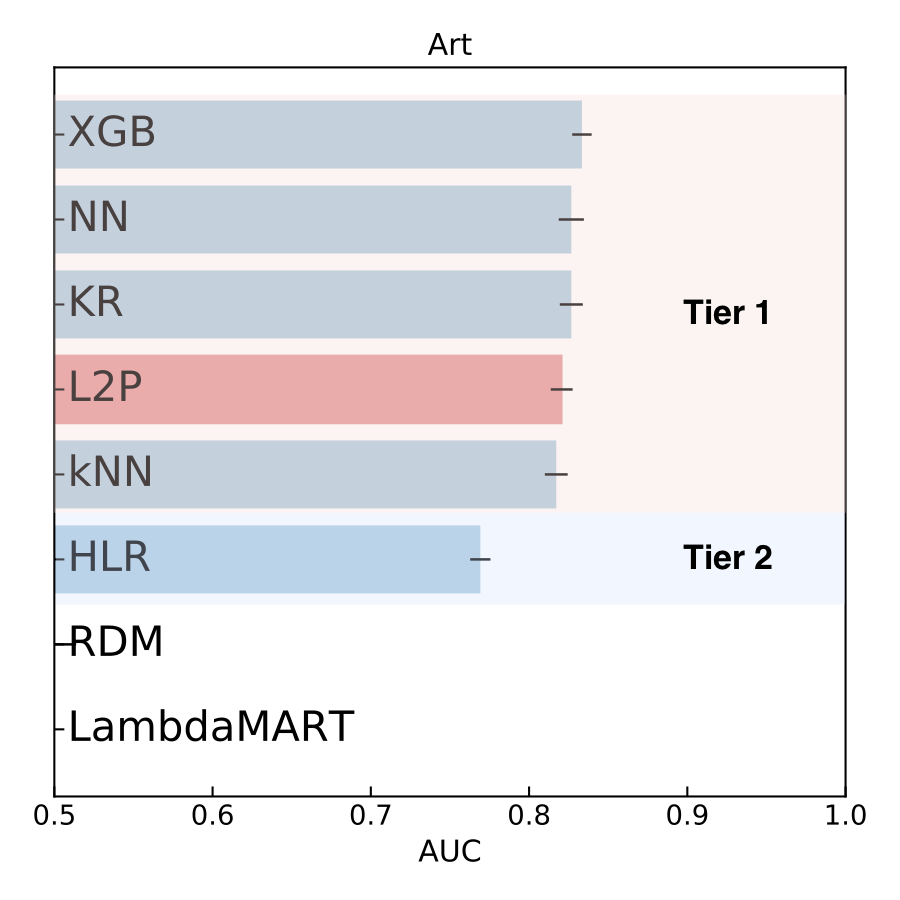} \\ 
  \includegraphics[height=0.295\linewidth]{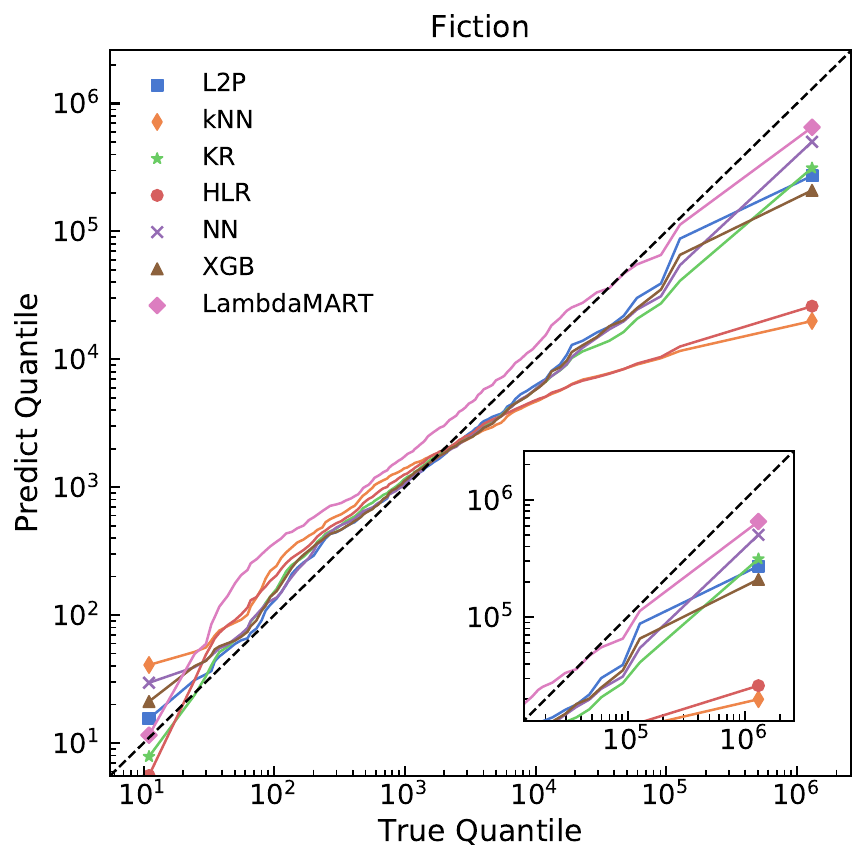}
  \includegraphics[height=0.30\linewidth]{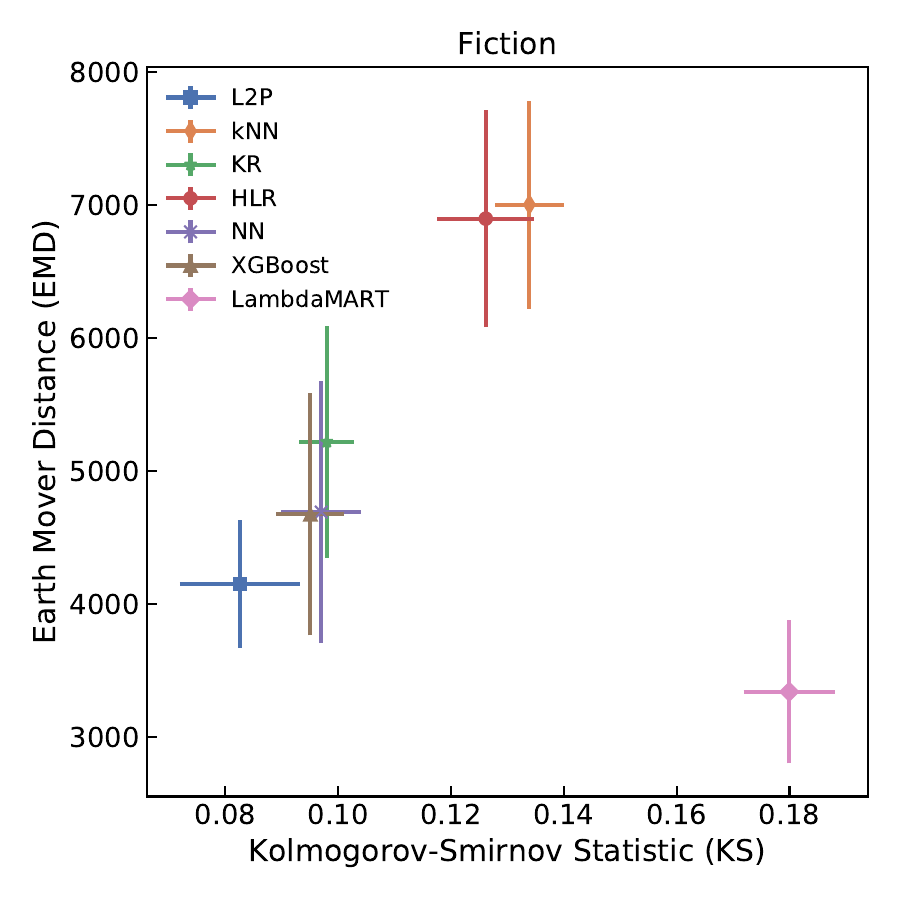}
  \includegraphics[height=0.30\linewidth]{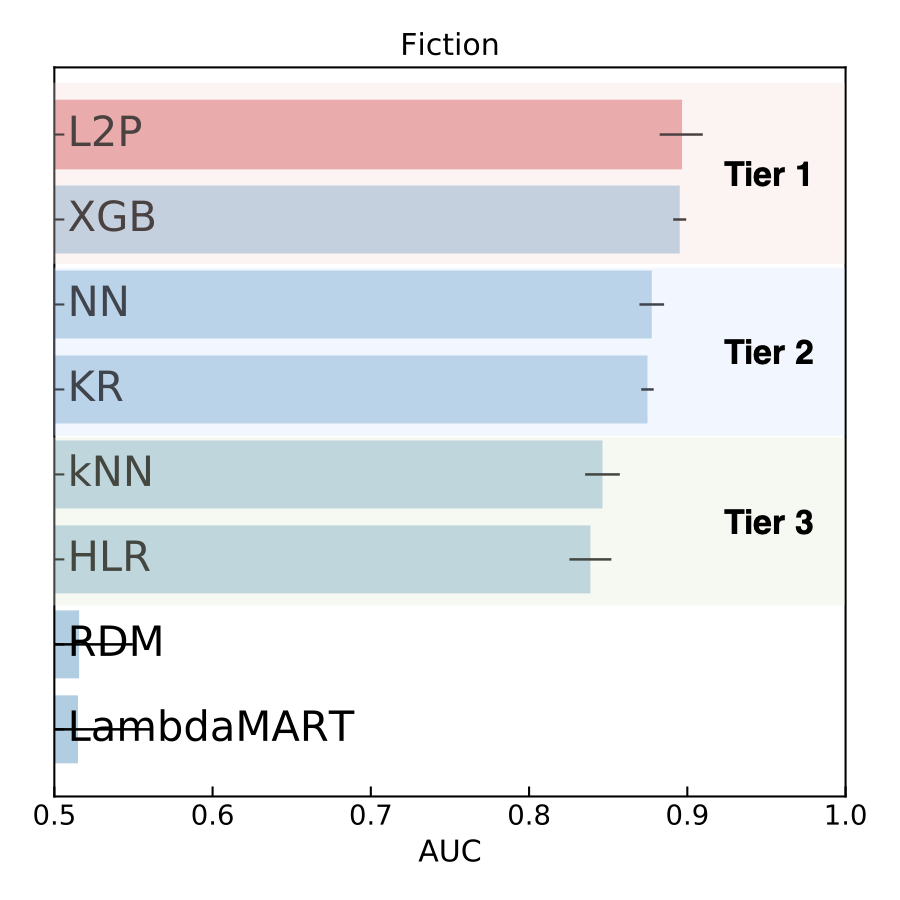} \\

  
\caption{\textbf{Experimental results}. We compare \texttt{L2P}'s performance against 6 other methods (discussed in Section~\ref{sec:othermethods}) across 3 datasets (see Table~\ref{tab:datasets}). 
\texttt{L2P}'s objective is to accurately reproduce the distribution of an outcome variable \underline{and} accurately predict its value. \textbf{Q-Q plots} (left column) show how predicted and actual quantiles align. \texttt{L2P} reproduces underlying distributions that are accurate at both lower and higher quantiles compared to the other methods. \textbf{Kolmogorov-Smirnov (KS) statistics} and \textbf{Earth mover distance (EMD)} (middle column) measures distance between predicted and actual distributions. The lower the values for KS and EMD, the better. \texttt{L2P} consistently outperforms the other methods. \textbf{AUC} (right column) measures prediction performance. We group methods to tiers based on the mean and standard deviation of the score. \texttt{L2P} achieves top tier performance all datasets.
In summary, the only approach that achieves top performance on both reproducing the outcome variable distribution and accurate predictions  across the datasets is \texttt{L2P}.}
  \label{fig:big_figure}
\end{figure*}

\subsubsection{Performance comparison study} We compare the methods using different measurement to showcase that \texttt{L2P} can accurately estimate the heavy-tailed distribution and predict the value of the heavy-tailed variable.

\textbf{Q-Q plot.} In left column in Fig.~\ref{fig:big_figure} shows the Q-Q plot of predicted outcomes. In all datasets 
we see deviation at the high-end (where ``big and rare'' instances reside). \texttt{L2P} is among the top 3 methods that produce the smallest deviation at the high end for all datasets. LambdaMART is also competitive in producing small deviation, but it produces larger deviations at the low end than \texttt{L2P}.

\textbf{KS and EMD.} The second column of Fig. \ref{fig:big_figure} presents the KS and EMD scores. Outcomes of \texttt{L2P} leads to the smallest KS statistics and second lowest EMD for almost all datasets. LambdaMART shows an advantage on minimizing EMD; however, we will show in AUC score comparison that it is not a preferable method.

\textbf{AUC.} The third column of Fig.~\ref{fig:big_figure} shows the AUC score of various methods on different datasets. \texttt{L2P} achieves top tier performance on nonfiction, art and fiction dataset, which all have high kurtosis values.
 XGB and NN have competitive AUCs to \texttt{L2P} but are bad at reproducing the true distributions as shown previously. We notice that LambdaMART, which has good performance EMD, has very low AUC score, indicating its inability to have accurate prediction on each instance. 
 
\begin{figure}[h!]
    \centering
    \includegraphics[width=0.4\columnwidth]{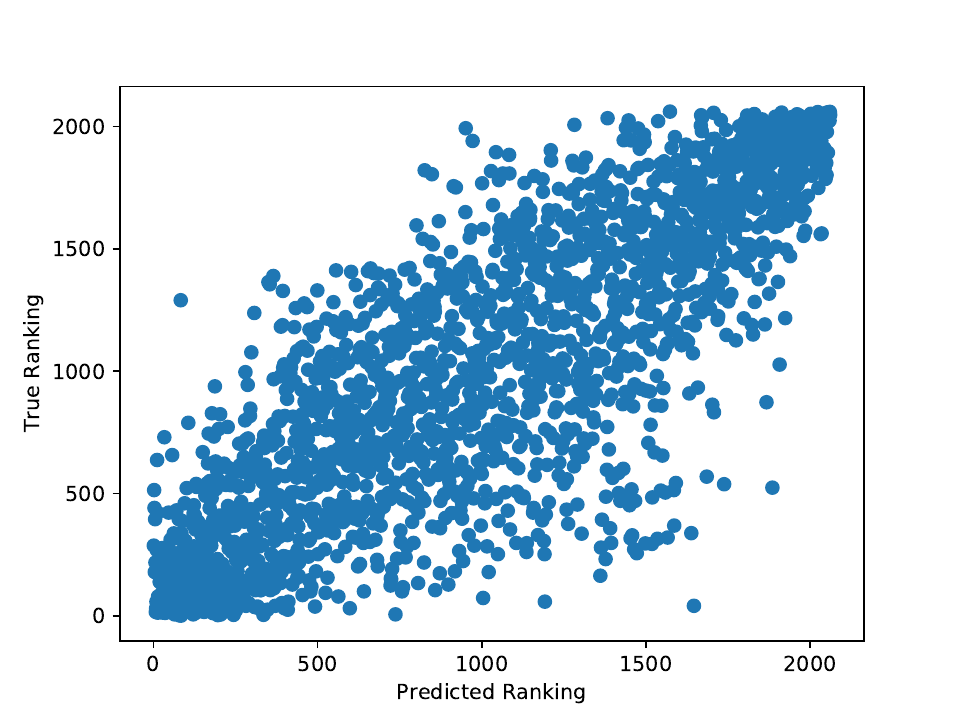}\\
    \caption{\textbf{Ranking performance for LambdaMART on the fiction dataset.}  LambdaMART is trained and tested on the full fiction dataset. We observe that even when trained on the same dataset, LambdaMART's ranking output is not accurate. While it exhibits general trends around the 45-degree line, the neighborhood ranking is not restored, which is the root cause of why LambdaMART fails to achieve good performance on predictions of new instances.}
    \label{fig:lambdamart-ranking}
\end{figure}

 We further investigate the possible reasons of the unsatisfying prediction of LambdaMART. Figure~\ref{fig:lambdamart-ranking} shows the ranking results for LambdaMART on the fiction dataset, where we trained and tested on the full dataset. 
 We found that LamdaMART did not correctly rank items that are close to each other. This failure of preserving the rank among training instances is the root cause of why LambdaMART fails to achieve good performance on the new instances prediction.

\textbf{Takeaway message.} With comprehensive consideration of all three evaluations, we can see that \texttt{L2P} is the best method in both reproducing the underlying heavy-tailed distribution and providing accurate predictions. 

\subsubsection{Robustness analysis}

In the placing phase of \texttt{L2P} there are two stages: 1) Stage I: obtaining pairwise relationships between test instance and training instances and 2) Stage II: placing the test instance and obtaining the prediction through voting. Previously we showed that voting itself is a maximum likelihood estimation, therefore the performance of \texttt{L2P} is highly depends on the performance of pairwise relationship learning. Here, we investigate the robustness w.r.t.~classification error of pairwise relationships.

To quantify the error tolerance of ``voting'' and estimation for a new instance, we conduct a set of experiments where we introduce errors on predicting pairwise relationships. 
We simulate the pairwise relationship error with two mechanisms: (i) \texttt{random error}: constant probability $p=p_c$ flips the label for each pair, (ii) \texttt{distance-dependent error}: probability of error is proportional to the true ranking percentile difference between items; here we use the percentile of the ranking because the sizes of the datasets vary. We define the flipping probability as $p_{ij}=e^{-\alpha |r_i-r_j|}$, assuming it would be easier to learn the pairwise relationship for items that are further away.  This is observed in our experiments as well. For example, in nonfiction data, we notice that more than 48\% of the pairwise relationship error occurs in item pairs that have a ranking percentile difference smaller than 10. We can control the rate of errors introduced by the two mechanisms by tuning $p_c$ or $\alpha$.

\begin{figure}[t!]
    \centering
    \includegraphics[width=0.4\columnwidth]{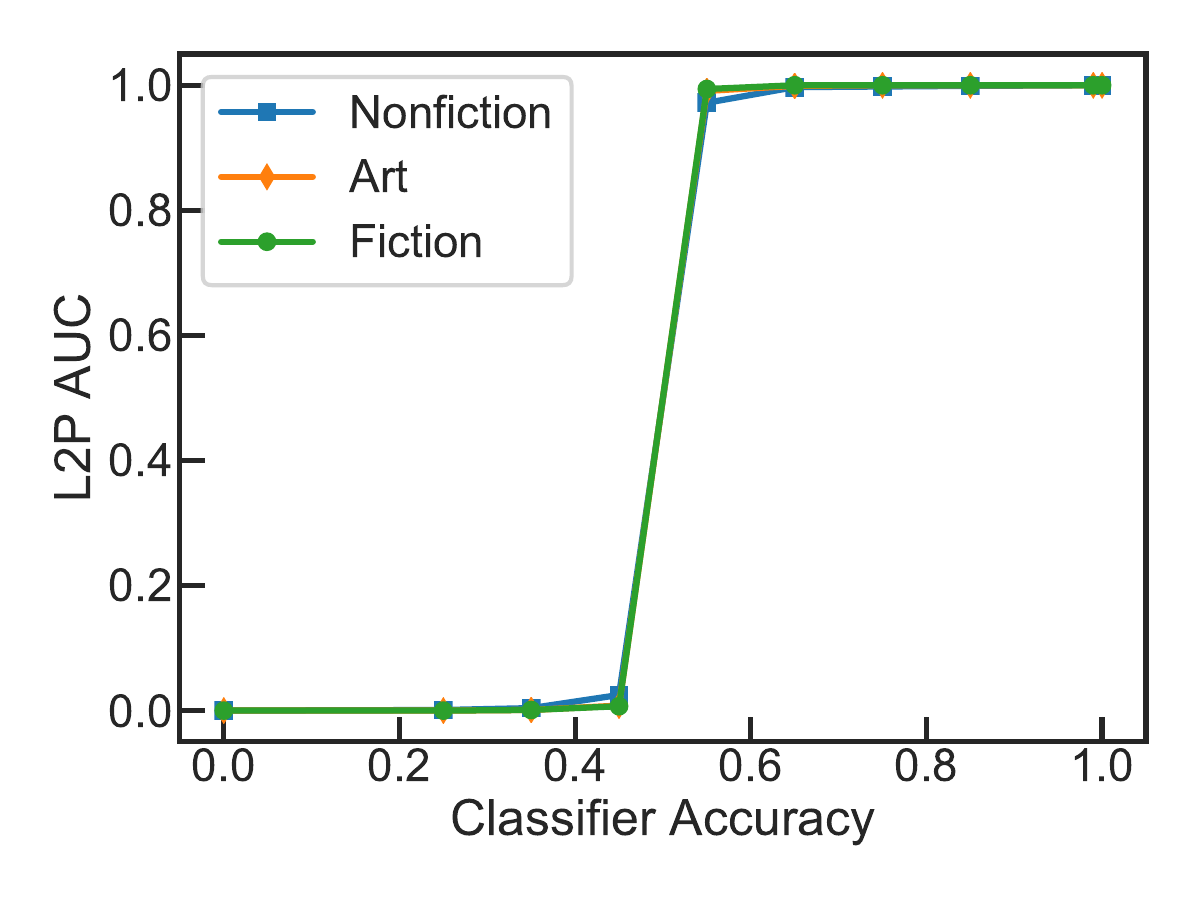}
    \includegraphics[width=0.4\columnwidth]{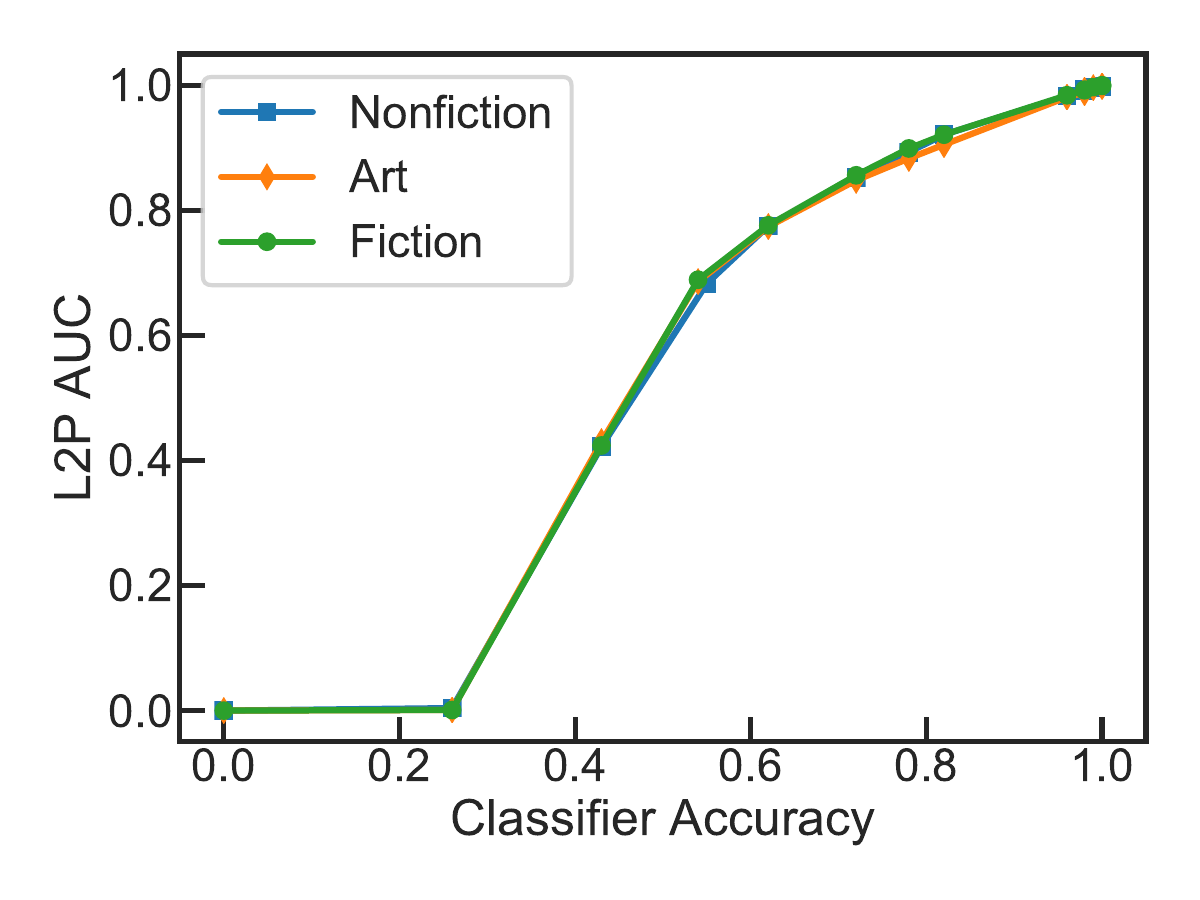}\\
    \caption{\textbf{\texttt{L2P}'s Robustness} \texttt{L2P}'s AUC scores after random errors (a) or distance-dependent errors (b) are \\introduced to \texttt{L2P}'s classifier for pairwise relationships. We observe significantly high tolerance towards random error and gradual degradation in \texttt{L2P}'s overall performance with distance-dependent errors.
    }
    \label{fig:robustness-analysis}
\end{figure}

In Figure~\ref{fig:robustness-analysis}, we present the overall performance (AUC) of \texttt{L2P} when various degrees of errors are introduced to the system in the pairwise relationships. First thing to notice is that if the pairwise relationship has no error (see left panel of Figure~\ref{fig:robustness-analysis} when classifier accuracy is 1), \texttt{L2P} has an accurate prediction, showing that the performance of the voting stage is only influenced by the quality of the pairwise relationships learned by the model. Moreover, the voting stage can actually compensate errors in pairwise relationships. We observe that error tolerance is significantly high towards random error: performance of \texttt{L2P} is stable until more than 45\% of the pairwise relationships are mistaken. For distance-dependent mechanism to simulate errors, we observe robust performance that 30\% error in stage I predictions resulting just 20\% reduction of the overall performance. 

\section{Model Interpretability}
As mentioned earlier, one of the advantages of  \texttt{L2P}'s methodology is its interpretability. 
Let us present the nonfiction book \texttt{Why not me?} by Mindy Kaling as an example. \texttt{L2P} prediction is about 218,000 copies while the actual sales is about 230,000. 
The key features explaining the success of this particular book are the author's popularity and the previous sales of author -- 6,228,182 pageviews and about 638,000 copies, respectively. However, performance of neural network leads to significant under-prediction of 16,000 sales and it's not clear why. \texttt{L2P} places \texttt{Why not me?} between \texttt{Selp-Helf} by Miranda Sings and \texttt{Big Magic} by Elizabeth Gilbert. \texttt{Selp-Helf} has the author popularity as 1,390,000 and the author has no prior publishing history, while \texttt{Big Magic} has the author popularity as 1,596,000 and previous sales as 6,954,000. We see that \texttt{Why not me?} has higher author popularity than \texttt{Big Magic} and \texttt{Selp-Helf}, but since it has a lower publishing history than \texttt{Big Magic}, \texttt{L2P} places it between these two books.

We also want to demonstrate an example case where \texttt{L2P} fails to achieve accurate prediction. The nonfiction book \texttt{The Best Loved Poems of Jacqueline Kennedy Onasis} by Caroline Kennedy under \texttt{Grand Central Publishing}, with claimed publication year 2015, is predicted to sell 53,000 copies while the actual sales is 180 copies in the dataset. After an extensive analysis, it turns out that the book was initially published in 2001 and was a New York Times bestseller, which \texttt{L2P} captures its potential and predict high sales.  Therefore this incorrect prediction is rooted in data error and our overprediction can be attributed to the initial editions performance as being a best-seller. Neural network predicts 7,150 copies, though closer to the actual sale 180. 

\section{Related Work and Discussion}

\begin{table*}[th!]
    \centering
    \begin{tabular}{|p{4cm}|c|c|c|c|c|c|c|}
        \hline
        \textbf{Objectives} & \textbf{L2P} & \textbf{kNN} & \textbf{KR} & \textbf{HLR} & \textbf{NN} & \textbf{XGB} & \textbf{LambdaMART} \\
        \hline
        Reproducing heavy-tailness & \cmark & \xmark & \xmark & \xmark & \cmark & \xmark & \cmark\\
    \hline 
        Accurate predictions & \cmark & \xmark & \cmark & \xmark & \cmark & \cmark & \xmark \\
        \hline
        Interpretable model & \cmark & \cmark & \xmark & \cmark & \xmark & \xmark & \cmark\\
        \hline
    \end{tabular}
    \caption{\textbf{Result summary.} We summarize performance of models on different objectives w.r.t.~whether they are in the top (\cmark) or bottom (\xmark) group. \texttt{L2P} satisfies all the objectives while the competing methods do not.} 
    \label{tab:summary}
\end{table*}



\noindent\textbf{Learning-to-rank methodologies.} Learning-to-rank is the application of machine learning in the construction of ranking models for information retrieval systems. In learning-to-rank, training data consists of lists of items with some partial order specified between items in each list and the ranking model's purpose is to rank, i.e. produce a permutation of items in new, unseen lists. 

Although \texttt{L2P} is designed for predicting heavy-tailed outcomes, which is different from learning-to-rank, methodological contributions show some parallels with the existing ranking algorithms. Cohen et al.~\cite{cohen1998learning} proposed a two-phase approach that learns from preference judgments and subsequently combines multiple judgments to learn a ranked list of instances. Similarly, RankSVM~\cite{joachims2002optimizing} is a two-phase approach that translates learning weights for ranking functions into SVM classification. Both of these approaches have complexity $O(n^2)$, which is computationally expensive.

In experiment practice, we found that learning-to-rank is not satisfying to predict heavy-tailed distributed outcomes. Since the learning-to-rank algorithm cannot guarantee to recover the rankings of the training instances completely, it will end up producing worse prediction on the new instance.



\noindent\textbf{Heavy-tailed regression.}
Regression problems are known to suffer from under-predicting rare instances~\cite{king2001logistic}. 
Approaches were proposed to correct fitting models consider prior correction that introduces terms capturing a fraction of rare events in the observations and weighting the data to compensate for differences~\cite{maalouf2018logistic,schubach2017imbalance}. 
Hsu and Sabato~\cite{hsu2014heavy} proposed a methodology for linear regression with possibly heavy-tailed responses. They split data into multiple pieces, repeat the estimation process several times, and select the estimators based on their performance. They analytically prove that their method can perform reasonably well on heavy-tailed datasets. Quantile regression related approaches are proposed as well. Wang \texttt{et al.}~\cite{wang2012estimation} proposed estimating the intermediate conditional quantiles using conventional quantile regression and extrapolating these estimates to capture the behavior at the tail of the distribution. Robust Regression for Asymmetric Tails (RRAT)~\cite{takeuchi2002robust} was proposed to address the problem of asymmetric noise distribution by using conditional quantile estimators. Zhang and Zhou~\cite{zhang2018ell_1} considered linear regression with heavy-tail distributions and showed that using $l1$ loss with truncated minimization can have advantages over $l2$ loss. Like all truncated based approaches, their method requires prior knowledge of distributional properties. However, none of these regression techniques can capture non-linear decision boundaries.

\noindent\textbf{Ordinal Regression.} The idea behind \texttt{L2P} methodology is similar to an ordinal regression where each training instance is mapped to an ordinal scale. Previous research has explored ordinal regression using binary classification~\cite{li2007ordinal}. The contribution of \texttt{L2P} is that it transforms the prediction of heavy-tailed outcomes to ordinal regression using pairwise-relationship classification followed by an MLE-based voting method.  By doing this two-phase approach, \texttt{L2P} is able to reproduce the distribution of the outcome variable and provide accurate predictions for the outcome variable.  Combining these two tasks leads consistently to better performance all-around.

\noindent\textbf{Imbalance Learning}
Data imbalance, as a common issue in machine learning, has been widely studied, especially in classification space. In~\cite{branco2016survey}, the problem of imbalance learning is defined as instances have different importance value based on user preference. There are in generally three categories of methods tackling this problem: data pre-processing~\cite{chawla2002smote,he2008adasyn}, special-purpose learning methods~\cite{maloof2003learning,torgo2007utility} and prediction post-processing~\cite{bansal2008tuning,sinha2004evaluating}. However, one should notice that learning heavy-tailed distributed attributes is different from imbalance learning. In most imbalance learning, there is a majority group and a minority group, but within group items are mostly homogeneous. However in heavy-tailed distribution, there is no clear cut to define majority/minority group and even if forcing a threshold to form majority/minority group, within each group, the distribution is still heavy-tailed. Additionally, one need to choose a pre-defined relevance function for a lot of methods designed in this space.


\noindent\textbf{Efficient algorithm for pairwise learning.} 
Qian et al. proposed using a two-step hashing framework to retrieve relevant instances and nominate pairs whose ranking is uncertain~\cite{qian2013fast}. Other approaches to efficiently searching for similar pairs and approximately learning pairwise distances are proposed in the literature for information retrieval and image search~\cite{chechik2009online,kulis2009fast,liu2016deep}. \texttt{L2P} can use any robust method that learns pairwise preferences for its pairwise relationships learning.

\section{Conclusions}

We presented the \texttt{L2P} algorithm, which satisfies three desired objectives consistently: (1) modeling heavy-tail distribution of an outcome variable, (2) accurately making predictions for the heavy-tailed outcome variable, and (3) producing an interpretable ML algorithm as summarized in Table~\ref{tab:summary}.
Through learning pairwise relationships following by an MLE-based voting method, \texttt{L2P} preserves the heavy-tailed nature of the outcome variables and avoids under-prediction of rare instances. We observed the following:

\begin{enumerate}
\item \texttt{L2P} accurately reproduces the heavy-tailed distribution of the outcome variable and accurately predicts of that variable.  Our experimental study, which included 6 competing methods and 3 datasets, demonstrates that \texttt{L2P} consistently outperforms other methods across various performance measures, including accurate estimation of both lower and higher quantiles of the outcome variable distribution, lower Kolmogorov-Smirnov statistic and Earth Mover Distance, and higher AUC.
 
\item \texttt{L2P}'s performance is robust when errors are introduced in the pairwise-relationship classifier. Under random error setting, \texttt{L2P} can achieve almost perfect performance up to 45\% error in pairwise relationship predictions; and under distance-dependent error setting, \texttt{L2P} has an accuracy drop of only 20\% with 30\% pairwise-relationship error.

\item  \texttt{L2P} is an interpretable approach and it provides prediction context. \texttt{L2P} allows one to investigate each prediction by comparing with neighboring instances and their corresponding feature values to obtain more context on the outcome. This is highly important to practitioners such as book publishers, where executives need reasons before making a huge investment.


\end{enumerate}





\bibliographystyle{unsrt}  
\bibliography{ms}
\end{document}


\maketitle

\section{Supplementary Material}
\subsection{Reproducibility}
The code for Python implementation of \texttt{L2P} method is freely available from \texttt{\url{https://github.com/xindi-dumbledore/L2P}}. Codes for baseline methods for the experiments are also included.

\subsection{Feature Description of Art Dataset}
Table~\ref{tab:art-feature} lists the features for the art dataset and its descriptions. The data includes \emph{exhibition grade}, which ranges from A (the top grade) to D (the bottom grade). To calculate average grade, we use the following assignments: A = 4, B = 3, C = 2 and D = 1.

\subsection{Parameters Selected for Different Baseline Methods and Datasets}
Table~\ref{tab:tuning-parameter} lists the parameters used for each baseline methods across different dataset. We performed a grid search on the parameters for each baseline method and select the one that produced the best performance.

\begin{table}[ht!]
\centering
\footnotesize
\begin{tabular}{|c|c|}
\hline
\textbf{Feature Name}                  & \textbf{Description}\\ \hline
previous\_shows\_count        & Number of exhibitions prior\\ \hline
previous\_A\_shows\_count     & Number of grade A exhibitions prior\\ \hline
previous\_B\_shows\_count     & Number of grade B exhibitions prior\\ \hline
previous\_C\_shows\_count     & Number of grade C exhibitions prior\\ \hline
previous\_D\_shows\_count     & Number of grade D exhibitions prior\\ \hline
previous\_shows\_avg\_grade   & average grade of previous exhibitions\\ \hline
previous\_sold\_item\_num     & Number of art pieces sold prior\\ \hline
previous\_median\_price       & Median price of art pieces sold prior\\ \hline
previous\_90percentile\_price & 90th percentile price of art pieces sold prior\\ \hline
previous\_75percentile\_price & 75th percentile price of art pieces sold prior\\ \hline
previous\_25percentile\_price & 25th percentile price of art pieces sold prior\\ \hline
previous\_10percentile\_price & 10th percentile price of art pieces sold prior\\ \hline
previous\_std\_price          & standard deviation of price of art pieces sold prior \\ \hline
show\_career\_length          & Number of years since first exhibition \\ \hline
sale\_career\_length          & Number of years since first auction sale \\ \hline
medium\_median                & Median sold price of art pieces of the same medium \\ \hline
medium\_percentile90          & 90th percentile of price of art pieces of the same medium \\ \hline
medium\_percentile75          & 75th percentile of price of art pieces of the same medium \\ \hline
medium\_percentile25          & 25th percentile of price of art pieces of the same medium \\ \hline
medium\_percentile10          & 10th percentile of price of art pieces of the same medium \\ \hline
medium\_std                   & standard deviation of price of art pieces of the same medium \\ \hline
\end{tabular}
\caption{\textbf{Feature table for art dataset.}\label{tab:art-feature}}
\end{table}

\begin{table}[ht!]
\centering
\footnotesize
\begin{tabular}{|l|l|p{1.5cm}|l|l|p{2cm}|p{2cm}|}
        \hline
        Data & kNN & KR & HLR & NN & XGB & LambdaMART \\
        \hline
        Nonfiction & 21 & $\alpha = 10^{-5}$, $\gamma = 5\times 10^{-5}$ & $\lambda = 5$ & (5, 10, 10) & max\_depth = 3, n\_estimator = 500, learning\_rate = 0.05 & max\_depth = 5, n\_estimator = 500, learning\_rate = 0.05 \\
    \hline 
        Art & 24 & $\alpha = 10^{-5}$, $\gamma = 5\times 10^{-5}$ & $\lambda = 0.1$ &(5, 5, 5) & max\_depth = 3, n\_estimator = 300, learning\_rate = 0.05 & max\_depth = 5, n\_estimator = 100, learning\_rate = 0.05 \\
        \hline
        Fiction & 20 & $\alpha = 5\times 10^{-3}$, $\gamma = 10^{-3}$ & $\lambda = 50$ & (5, 10, 10) & max\_depth = 3, n\_estimator = 300, learning\_rate = 0.05 & max\_depth = 5, n\_estimator = 500, learning\_rate = 0.05\\
        \hline
\end{tabular}
    \caption{\textbf{Model parameters of baseline methods for different dataset.}\label{tab:tuning-parameter}}
\end{table}